\documentclass[letterpaper, 10 pt, conference]{ieeeconf}  % Comment this line out if you need a4paper

\IEEEoverridecommandlockouts                              % This command is only needed if 
\usepackage{graphics} % for pdf, bitmapped graphics files
\usepackage{epsfig} % for postscript graphics files
\usepackage{mathptmx} % assumes new font selection scheme installed
\usepackage{times} % assumes new font selection scheme installed
\usepackage{amsmath} % assumes amsmath package installed
\usepackage{amssymb}  % assumes amsmath package installed
\usepackage{multirow}
\usepackage{bbding}
\usepackage{url}
\usepackage{array} 
 
\usepackage{bm}
\usepackage{algorithm}
\usepackage{algorithmic}

\title{\LARGE \bf
A Neuromorphic Reinforcement Learning Framework for Efficient Pathfinding in Robotic Mobile Fulfillment Systems
}

\author{Junzhe Xu$^{1,2}$, Zecui Zeng$^{2,*}$, Lusong Li$^{2}$, Yuetong Fang$^{1,\dagger}$, Renjing Xu$^{1,\dagger}$% <-this % stops a space
\thanks{This work was supported by the National Key Research and Development Program of China (No. 2024YFE0211000).}% <-this % stops a space
\thanks{$^{1}$ The Hong Kong University of Science and Technology (Guangzhou), Guangzhou, China.}%
\thanks{$^{2}$ JD Explore Academy, Beijing, China.}% 
\thanks{$^\dagger$ Corresponding authors.}%
\thanks{$^*$ Project Leader.}
}

\begin{document}

\maketitle
\thispagestyle{empty}
\pagestyle{empty}

%%%%%%%%%%%%%%%%%%%%%%%%%%%%%%%%%%%%%%%%%%%%%%%%%%%%%%%%%%%%%%%%%%%%%%%%%%%%%%%%
\begin{abstract}
Dynamic environmental changes, confined workspaces, and stringent real-time constraints make pathfinding in Robotic Mobile Fulfillment Systems (RMFS) a challenging problem for conventional search- and rule-based methods, which typically suffer from high computational complexity and long decision latency. While reinforcement learning (RL) has emerged as a powerful alternative, deploying learned policies with extreme energy efficiency on resource-constrained hardware remains an open challenge. We present SDQN-RMFS, an end-to-end framework that achieves high-fidelity deployment of an RL-trained policy from a full-precision artificial neural network (ANN) through to a neuromorphic chip. By computing only when triggered by sparse events, this framework unlocks ultra-low-power RMFS pathfinding. Our full-stack pipeline operates as follows: an ANN policy is first efficiently trained via a collision-allowing strategy to densify informative trajectories, and then converted into a spiking neural network (SNN) via a hard-label knowledge distillation approach. This effectively addresses the output distribution mismatch, preserving policy capability across the ANN-to-SNN pipeline while substantially reducing inference latency. Hardware experiments demonstrate up to 11,281$\times$ energy savings and a nearly two-fold reduction in latency compared to a high-performance GPU baseline, while maintaining decision quality on par with the original trained policy. These results establish physical neuromorphic inference as a practical and energy-sustainable pathway for large-scale RMFS operations.

% We present SDQN-RMFS, a framework that, for the first time, achieves lossless end-to-end deployment of an RL-trained policy from a full-precision artificial neural network (ANN) through to a neuromorphic chip that computes only when triggered by sparse events, unlocking ultra-low-power RMFS pathfinding. The core breakthrough lies in our full-pipeline conversion and deployment methodology: an ANN policy is first efficiently trained via a collision-allowing strategy that densifies informative state–action trajectories; it is then converted to a spiking neural network (SNN) via the hard-label knowledge distillation approach that enforces functional equivalence across the full ANN-to-SNN-to-chip pipeline while substantially reducing inference latency. Hardware experiments demonstrate up to 11,281$\times$ energy savings and a nearly two-fold reduction in inference latency compared to GPU baselines, while maintaining decision quality on par with the original trained policy. These results establish neuromorphic deployment as a practical pathway for large-scale, energy-sustainable RMFS operations.

\end{abstract}

\section{Introduction}
Robotic Mobile Fulfillment Systems (RMFSs) have transformed modern warehouse logistics by deploying fleets of Automated Guided Vehicles (AGVs) to autonomously transport inventory pods to picking stations on demand, dramatically reducing reliance on manual labor and improving order throughput \cite{da2021robotic}. Large-scale RMFSs involve hundreds to thousands of AGVs operating simultaneously in shared and resource-constrained spaces, making efficient and safe autonomous navigation a critical operational bottleneck~\cite{sartoretti2019primal,damani2021primal}. AGVs must continuously plan collision-free routes through narrow passages and dynamically changing obstacles (including other moving AGVs) under strict real-time latency requirements. Classical graph search methods such as Dijkstra \cite{dijkstra2022a} and A* \cite{hart1968formal} are fundamentally ill-suited to this setting: designed for static, single-agent environments, they require complete replanning upon any state change, incurring prohibitive computational overhead as the number of agents and environment complexity grow. Extensions to the multi-agent domain~\cite{bolu2019path,chen2021integrated} partially address collision avoidance but retain worst-case exponential complexity with respect to agent count, rendering them computationally prohibitive for highly dynamic, large-scale RMFS deployments.

Researchers have increasingly incorporated reinforcement learning (RL) into AGV pathfinding as a principled alternative to classical search methods~\cite{mnih2015human}. By enabling agents to iteratively refine their decision policies through environmental interaction, RL naturally accommodates the dynamic, rule-free nature of warehouse navigation~\cite{kamoshida2017acquisition}. Recent advances in deep reinforcement learning (DRL) have further extended its applicability, achieving notable success in multi-agent settings \cite{arulkumaran2017deep,damani2021primal,luo2023guiding}. Nevertheless, DRL inference relies on synchronous, dense floating-point operations. This remains computationally demanding and energy-intensive, fundamentally limiting its viability in mobile AGVs operating under strict size, weight, and power constraints at the edge~\cite{tang2020reinforcement,zhang2025toward}.

Spiking neural networks (SNNs) offer a compelling solution to this efficiency bottleneck by emulating the event-driven spike transmission mechanisms of biological neurons \cite{maass1997networks}. When deployed on neuromorphic hardware, SNNs perform computation only upon the arrival of sparse asynchronous spikes, yielding orders-of-magnitude reductions in energy consumption compared to conventional ANN inference on GPUs~\cite{davies2018loihi}. However, realizing this efficiency advantage within an RL framework remains non-trivial. The majority of existing RL-SNN studies operate entirely in simulation, stopping short of actual neuromorphic chip deployment and thus failing to substantiate practical energy savings~\cite{patel2019improved,tang2020reinforcement,zhang2025toward,salvatore2020neuro}. Moreover, conventional ANN-to-SNN conversion methods are designed for supervised classification tasks with sparse one-hot label distributions, and do not generalize well to the dense, continuous action-value distributions that characterize RL policies. This mismatch introduces non-trivial accuracy degradation upon conversion.

% On the other hand, SNNs offer a novel computational paradigm by simulating the spike transmission mechanisms of biological neurons \cite{maass1997networks}. Compared to conventional artificial neural networks (ANNs), SNNs deployed on neuromorphic chips can significantly reduce energy consumption during inference due to their event-driven and asynchronous computation characteristics\cite{davies2018loihi}. While SNNs have shown promise, integrating them into RL frameworks presents unique challenges \cite{patel2019improved,tang2020reinforcement,zhang2025toward}. The majority of existing RL-SNN studies remain at the simulation level, lacking actual neuromorphic chip deployment, thereby failing to demonstrate practical energy efficiency benefits. Furthermore, traditional ANN-to-SNN conversion methods are tailored for supervised learning with sparse one-hot distributions, which inherently conflict with the dense continuous action probability distributions required in RL.

Building on these insights, we propose SDQN-RMFS, a framework that systematically addresses each identified challenge along the full pipeline from RL policy training to real neuromorphic chip deployment for multi-AGV pathfinding in RMFSs. Rather than treating training, conversion, and deployment as independent sub-problems, SDQN-RMFS establishes a unified, end-to-end pathway that rigorously preserves policy fidelity at every stage while realizing the substantial energy advantages that neuromorphic hardware uniquely affords. The main contributions of this work are:
\begin{itemize}
\item We introduce SDQN-RMFS, a framework designed for RL-based multi-agent pathfinding in warehouse environments. To accelerate policy convergence, we devise a robust DQN architecture enabled by a collision-allowing training strategy.
\item We propose a hard-label knowledge distillation scheme for ANN-to-SNN conversion. This method aligns the continuous action-value distribution of RL with the discrete nature of SNNs, preserving high-fidelity decision capability while substantially reducing inference latency.
\item We perform offline physical deployment of the converted SNN directly onto the SPECK2E~\cite{richter2023speck} neuromorphic chip. Comprehensive hardware evaluations confirm up to 11,281$\times$ energy savings and reduced latency by approximately half relative to an NVIDIA RTX 4090 GPU baseline, substantiating the edge applicability of the framework.

% \item We introduce SDQN-RMFS, a framework for RL-based multi-AGV pathfinding in warehouse environments. To substantially accelerate policy convergence, we devise a robust DQN architecture enabled by the collision-allowing training strategy that promotes comprehensive coverage of the state–action trajectory space.
% \item We propose a hard-label knowledge distillation scheme for ANN-to-SNN conversion that enforces strict output correspondence between the SNN and its ANN counterpart, ensuring functional equivalence throughout the complete ANN-to-SNN-to-chip deployment chain while substantially reducing inference latency.
% \item We deploy the converted SNN directly onto the SPECK2E~\cite{richter2023speck} neuromorphic chip and conduct comprehensive hardware evaluations, confirming up to 11,281$\times$ energy savings relative to an NVIDIA RTX 4090 GPU and a reduction in inference latency by approximately half, substantiating the practical edge applicability of the proposed framework.
\end{itemize}

\section{Related Work}
\subsection{Pathfinding in RMFSs}
% To address the challenges of AGV pathfinding in the complex environments of RMFS, several studies have proposed search-based or rule-based path planning algorithms. \cite{bolu2019path} developed a collision-free path planning method based on the A* algorithm, enabling efficient operation of multiple robots across various scenarios. \cite{chen2021integrated} explored a hybrid path planning strategy that amalgamates traditional pathfinding algorithms with heuristic methods, designing a marginal-cost assigning heuristic and a meta-heuristic improvement strategy while concurrently addressing both task allocation and path planning issues within RMFSs. In response to the limitations inherent in traditional algorithms, researchers have begun exploring emerging algorithms. \cite{sartoretti2019primal} integrated distributed reinforcement learning with imitation learning from a centralized expert planner, demonstrating superior performance. \cite{luo2023guiding} incorporated the A* algorithm into a reinforcement learning framework to serve as a guiding strategy for agents, further enhancing the speed of agent training. We take this inspiration of combining RL with RMFSs and further improve it for more efficient pathfinding.

Traditional AGV pathfinding in Robotic Mobile Fulfillment Systems (RMFSs) primarily relies on search-based or rule-based algorithms. For instance, A*-based methods and hybrid heuristic strategies have been widely adopted to achieve collision-free navigation and concurrent task allocation \cite{bolu2019path, chen2021integrated}. While effective in controlled scenarios, these traditional algorithms often struggle with computational bottlenecks and scalability in highly dynamic and complex environments. Consequently, recent literature has increasingly shifted towards learning-based approaches. Distributed reinforcement learning, particularly when augmented with imitation learning or heuristic guidance like A*, has demonstrated superior adaptability and accelerated training speeds \cite{sartoretti2019primal, luo2023guiding}. Building upon this trajectory, our work leverages the robust decision-making capabilities of RL for complex RMFS dynamics, while further seeking breakthroughs in computational and energy efficiency.

\subsection{RL with SNN}\label{sec.rlsnn}
% Recent research suggests that integrating SNNs with RL frameworks can facilitate more efficient and biologically plausible decision-making processes. \cite{patel2019improved} was the first to employ SNNs for DRL tasks, demonstrating that SNNs exhibit greater robustness than traditional ANNs in the context of playing Atari games. \cite{tang2020reinforcement} developed a hybrid framework, termed SDDPG, which consists of a spiking actor network and a deep critic network. This framework successfully deployed the jointly trained SNN onto neuromorphic computing chips, enabling a mobile robot to navigate through a mapless environment, thereby offering an efficient and energy-conserving solution. \cite{tan2021strategy} implemented a conversion from ANN to SNN to facilitate the integration of DQNs with SNNs, while proposing a series of optimization measures to mitigate performance loss during the conversion process. Through extensive experimentation, \cite{tan2021strategy} validated both the feasibility and potential advantages of integrating RL with SNNs. However, this efficient and energy-saving approach combining RL with SNNs has not yet been validated and applied in the context of pathfinding tasks within warehousing and logistics scenarios.

The deployment of deep RL in real-world robotic systems is often constrained by significant energy consumption. SNNs have emerged as a biologically plausible and energy-efficient alternative. Pioneering studies have successfully integrated SNNs with RL frameworks, proving their superior robustness in discrete tasks like Atari games \cite{patel2019improved}. Subsequent research expanded this to continuous control and mapless navigation by developing hybrid spiking actor-critic architectures and optimizing ANN-to-SNN conversion techniques \cite{tang2020reinforcement, tan2021strategy}. These advances demonstrate the feasibility and energy-saving potential of SNN-based RL. However, a critical gap remains: existing RL-SNN works remain confined to simulation without physical deployment on neuromorphic hardware, leaving the promised energy efficiency gains empirically unsubstantiated.

%this highly efficient, neuromorphic RL paradigm has yet to be adapted and validated for large-scale, multi-agent pathfinding tasks within warehousing and logistics contexts.

\subsection{ANN-to-SNN Conversion}\label{sec.atos}

To obtain deep SNNs, ANN-to-SNN conversion is currently the mainstream approach as it effectively circumvents the non-differentiable optimization challenges inherent in spiking dynamics. Extensive efforts have focused on minimizing conversion loss and inference latency through techniques such as threshold balancing \cite{diehl2015fast}, Rate Norm Layers \cite{deng2021optimal}, advanced quantization functions \cite{bu2022optimal}, and layer-wise parameter calibration \cite{li2024error}. Despite these significant improvements, a fundamental limitation persists: these optimization techniques are predominantly tailored for supervised learning tasks (e.g., image classification), which target sparse, one-hot probability distributions. In stark contrast, RL intrinsically relies on precise continuous value estimation and dense action probability distributions. Directly applying existing conversion methods to RL introduces severe discrete quantization errors at low time-steps, inevitably leading to policy degradation. Addressing this exact mismatch between the discrete spike-based representations of SNNs and the continuous action-value inherent to RL policies requirements constitutes the central technical challenge motivating the present work.

\section{Preliminary}
\subsection{Robotic Mobile Fulfillment Systems}
\begin{figure}[t]
\centering
\includegraphics[width=0.9\columnwidth]{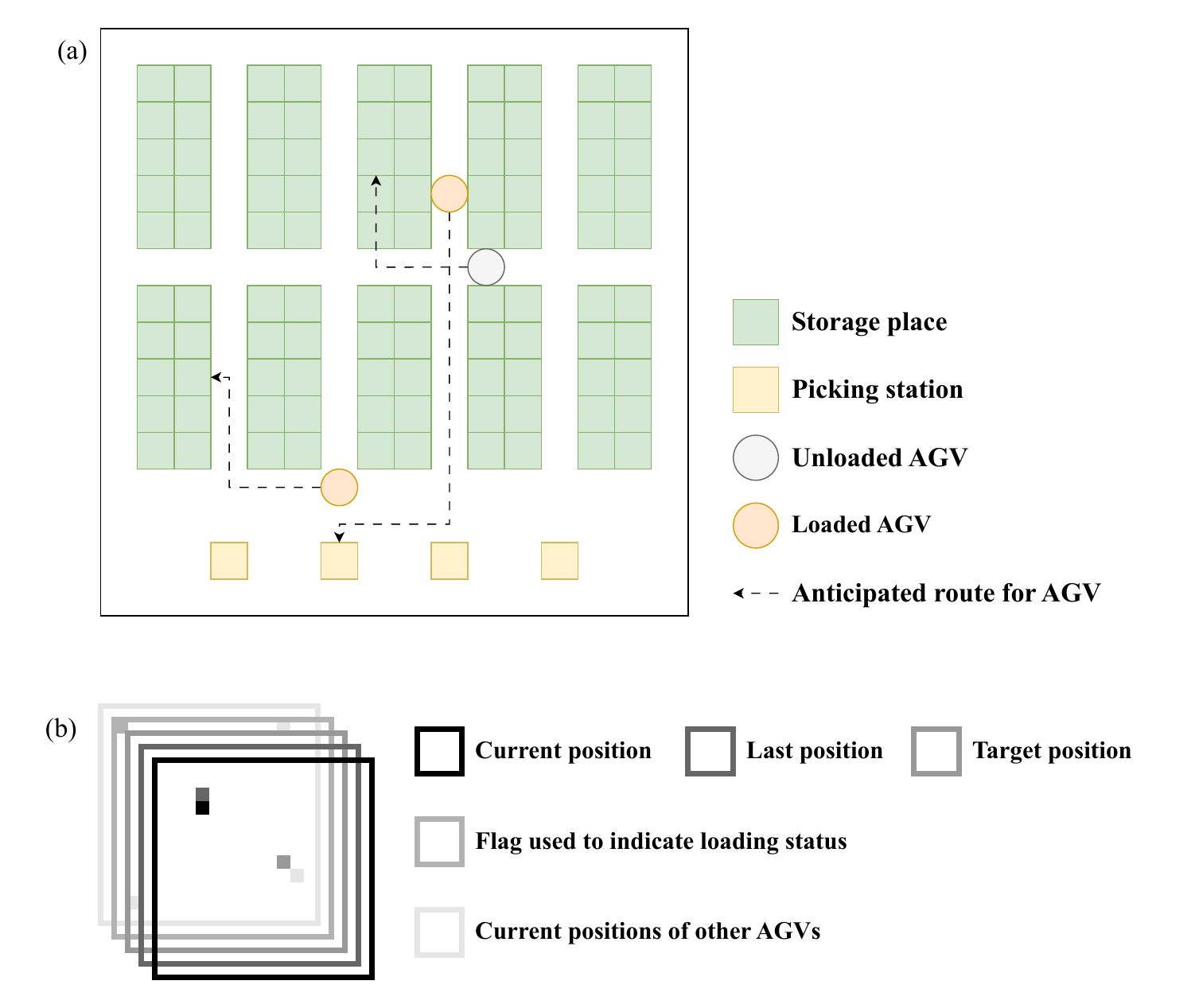}
\caption{(a) Sketch map of RMFS, which includes several AGVs with or without loading, storage places, and picking stations. The dashed arrows represent the anticipated routes for both loaded and unloaded AGVs as they navigate the warehouse. (b) The state input for DQN, which is represented by a set of maps that include the AGV's current position, last position, target position, loading status, and the current positions of other AGVs.}
\vspace{-0.4cm}
\label{fig.rmfs}
\end{figure}

In a typical RMFS environment (Figure \ref{fig.rmfs}a), AGVs execute a three-stage transport cycle: navigating unloaded to a storage area to load a pod, transporting the loaded pod to a picking station, and finally returning it to its original location before reverting to an unloaded state. Throughout this process, AGVs must strictly adhere to boundary constraints and collision avoidance regarding both storage units and other vehicles; moreover, a key operational advantage allows unloaded AGVs to traverse beneath storage pods to optimize pathing. Figure \ref{fig.rmfs}a offers a detailed representation of the optimal routes and operational states of the AGVs during these three stages.

\subsection{Spiking Neural Networks and IF Model}\label{sec.if}
Unlike traditional ANNs that communicate via continuous values, SNNs process information using discrete binary spikes, closely mimicking biological neural mechanisms. This event-driven nature allows SNNs to achieve remarkable energy efficiency when deployed on neuromorphic hardware.

To circumvent the Backpropagation Through Time (BPTT) overhead, we adopt the ANN-to-SNN conversion method. Similar to prior works, we utilize the integrate-and-fire (IF) spiking neuron model \cite{tan2021strategy,wang2023masked}. For an SNN with $L$ layers, the membrane potential $v^l(t)$ updates via a subtracting mechanism:
\begin{equation}
    v^l(t)=v^l(t-1)+z^l(t)-V_{th}\theta^l(t)\label{eq.snn}
\end{equation}
where $z^l(t)=W^l\theta^{l-1}(t)+b^l$ is the input current, $V_{th}$ is the firing threshold (typically set to 1), and $\theta^l(t)$ represents the output spike generated via the Heaviside step function when the potential exceeds the threshold.

\section{Method}
\subsection{Overview}\label{sec.overview}
\begin{figure*}[t]
\centering
\includegraphics[width=0.9\textwidth]{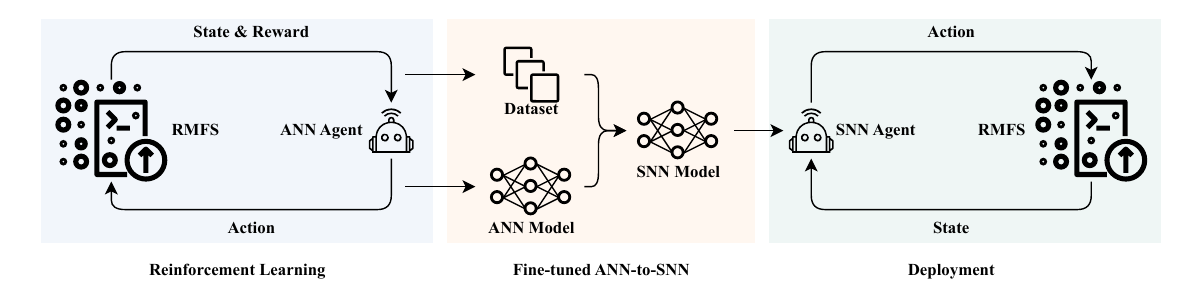}
\vspace{-0.4cm}
\caption{The three-stage process of the SDQN-RMFS framework. First, an ANN agent is trained using RL to interact with the RMFS, collecting state-action trajectories. Second, a fine-tuned ANN-to-SNN conversion is applied to transform the trained ANN model into an SNN model, preserving its capabilities. Finally, the SNN agent is deployed for efficient and energy-saving action inference in the RMFS.}
\vspace{-0.4cm}
\label{fig.overview}
\end{figure*}

We outline the workflow of the proposed SDQN-RMFS framework, as depicted in Figure \ref{fig.overview}. The pipeline consists of three primary stages. First, a high-performance ANN agent is trained via a fundamentally enhanced DQN through continuous interactions with the RMFS environment. This phase also facilitates the collection of comprehensive state-action trajectories. Second, utilizing the collected state-action trajectories, we apply a robust ANN-to-SNN conversion incorporating parameter scaling and knowledge distillation. Finally, the resulting SNN agent is deployed onto neuromorphic hardware, achieving highly efficient and energy-conserving pathfinding operations.

To strengthen training stability, we incorporate a target network alongside a prioritized experience replay mechanism, and adopt the Double DQN (DDQN) formulation to decouple action selection from action evaluation \cite{van2016deep}. For brevity, this enhanced agent is referred to simply as 'DQN' throughout the remainder of this paper.

% Figure 2 illustrates the SDQN-RMFS workflow: (1) ANN training via a collision-allowing strategy, (2) robust ANN-to-SNN conversion incorporating parameter scaling and knowledge distillation, and (3) physical deployment onto neuromorphic hardware. To strengthen training stability and mitigate Q-value overestimation [25], we utilize Double DQN (DDQN) with prioritized experience replay, simply referred to as 'DQN' hereafter.

%To enhance performance, we incorporate a target network and a prioritized experience replay mechanism. Specifically, our training process utilizes the Double DQN (DDQN) methodology, which decouples action selection from action evaluation to effectively mitigate the overestimation of Q-values often observed in standard Q-learning \cite{van2016deep}. For brevity, we refer to this enhanced model simply as 'DQN' throughout the remainder of this paper. This allows the DQN to effectively perform pathfinding in RMFS environments.

\subsection{Problem Formulation}\label{sec.problem}
We formulate this RMFS pathfinding task as a Markov Decision Process (MDP). The agent observes a state $s=\{pos_c, pos_l, pos_t, loaded, pos_o\}$, where $pos_c$, $pos_l$, and $pos_t$ denote the current, previous, and target positions of the AGV, respectively; $loaded$ indicates the operational status; and $pos_o$ tracks the positions of other AGVs. The action space $|a|=5$ consists of four cardinal movements and a stationary action (reducing to $|a|=4$ in single-agent scenarios).

Despite the theoretical potential of combining DQN with SNNs for this MDP, directly applying standard methods reveals three critical bottlenecks that our method must address:
\begin{itemize}
    \item Exploration Stagnation in Constrained Spaces: The RMFS environment is characterized by numerous narrow passages. Standard random exploration in DQN frequently leads to immediate collisions and task failures. This prevents the agent from securing positive rewards, causing premature learning cessation and severe training inefficiency.
    \item Insufficient Information Transmission at Low Time-steps: To achieve the desired ultra-low inference latency and energy efficiency on edge hardware, the SNN must operate at extremely low time-steps. However, due to the zero-initialized membrane potential ($v^l(0)=0$), deep spiking neurons suffer from delayed spike emission. This prevents the effective forward transmission of information within the limited time-step window, rendering the SNN dysfunctional.
    \item RL-SNN Output Distribution Mismatch: Standard ANN-to-SNN conversion assumes a sparse, one-hot output distribution typical of classification tasks. However, RL agents output continuous action values (Q-values) that exhibit a flat distribution. At the ultra-low time-steps required for edge hardware deployment, this mismatch introduces severe discrete quantization errors, inevitably leading to policy degradation post-conversion.
\end{itemize}
 
\subsection{SNN-friendly DQN with Collision-Allowing Strategy}
To address the inherent complexities of the RMFS and the exploration stagnation outlined in Section \ref{sec.problem}, we design an enhanced DQN framework customized for neuromorphic deployment.

\subsubsection{SNN-Friendly State Representation}\label{sec.rep}
As shown in Figure \ref{fig.rmfs}b, the inputs are natively binary, with only a sparse fraction of elements activated as 1. This inherently discrete representation is specifically designed to be SNN-friendly: the raw state matrices act directly as spatial spikes. By entirely bypassing the need for intermediate rate- or latency-coding conversions during neuromorphic deployment, this native spike input pipeline fundamentally guarantees ultra-low inference latency and minimizes energy consumption from the ground up.

\subsubsection{Reward Shaping}
In RL, reward drives both goal-learning and the exploration-exploitation balance. In RMFS environments, the agent receives a positive reward $R_{goal}$ for reaching the target and a negative reward $R_{fail}$ for collisions or exceeding boundaries. To discourage ineffective repetitive exploration, we implement a shaped reward $R_{s}$ based on the Manhattan distance $D$:
\begin{equation}
    R_{s}=
    \begin{cases}
    -0.5&\text{if}\ \  pos_l=pos_n\\
    0.1&\text{if}\ \  pos_l\not=pos_n\ \text{and}\ D(pos_n)<D(pos_c)\\
    -0.1&\text{if}\ \  pos_l\not=pos_n\ \text{and}\ D(pos_n)=D(pos_c)\\
    -0.2&\text{if}\ \  pos_l\not=pos_n\ \text{and}\ D(pos_n)>D(pos_c)\\
    \end{cases}
\end{equation}
where $pos_n$ and $pos_l$ represent the next and last positions of the AGV.

\subsubsection{Collision-Allowing Exploration}

In RMFS environments, narrow passages are ubiquitous and pose a far greater challenge than standard isolated obstacles. Traversing them demands continuous precision, whereas a single random exploratory move often results in an immediate collision and task failure, rapidly causing severe exploration stagnation. To fundamentally overcome this constraint, we introduce a collision-allowing strategy. During each training episode, we allow a certain number of collisions to occur, as detailed in Algorithm \ref{alg:collision_step}. Rather than immediately terminating an episode upon collision, the agent is penalized and reverted to its previous position, provided the collision count remains below the limit $CL$. This collision-allowing strategy aims to enable exploration within a certain range, thereby preventing learning stagnation resulting from frequent failures. As training progresses, the proportion of random exploration in the DQN will gradually decrease, ultimately resulting in the agent's decisions relying entirely on the DQN's output.

\begin{algorithm}[b!]
\caption{Collision-Allowing Exploration}
\label{alg:collision_step}
\begin{algorithmic}[1] 
\REQUIRE Current state $s_t$, Action $a_t$, Collision limit $CL$, Current collision count $c_{count}$
\STATE Execute action $a_t$, observe collision status $is\_collision$ and target status $is\_target$
\IF{$is\_collision$}
    \STATE $c_{count} \leftarrow c_{count} + 1$
    \IF{$c_{count} \leq CL$}
        \STATE $r_t \leftarrow R_{fail}$ \COMMENT{Penalize but allow continuation}
        \STATE $s_{t+1} \leftarrow s_t$ \COMMENT{Revert AGV to previous position}
        \STATE $done \leftarrow \text{False}$
    \ELSE
        \STATE $r_t \leftarrow R_{fail}$
        \STATE $done \leftarrow \text{True}$ \COMMENT{Exceed limit, terminate episode}
    \ENDIF
\ELSE
    \STATE Calculate shaped reward $R_s$
    \IF{$is\_target$}
        \STATE $r_t \leftarrow R_{goal}$
        \STATE $done \leftarrow \text{True}$
    \ELSE
        \STATE $r_t \leftarrow R_s$
        \STATE $done \leftarrow \text{False}$
    \ENDIF
    \STATE $s_{t+1} \leftarrow \text{true next state}$
\ENDIF
\RETURN $s_{t+1}, r_t, done, c_{count}$
\end{algorithmic}
\end{algorithm}

% In RMFS environments, narrow passages are ubiquitous and pose a far greater challenge than standard isolated obstacles. Traversing them demands continuous precision; a single random exploratory move often results in an immediate collision and task failure, rapidly causing severe exploration stagnation. To fundamentally overcome this constraint and enable the agent to discover viable paths, we introduce a collision-allowing strategy to make improvements to the training framework. During each training episode, we allow a certain number of collisions to occur. If the number of collisions does not exceed a predetermined collision limit $CL$, the colliding agent is returned to its last position and retains the negative reward, thereby continuing the learning process. This collision-allowing strategy aims to enable exploration within a certain range, thereby preventing learning stagnation resulting from frequent failures. As training progresses, the proportion of random exploration in the DQN will gradually decrease, ultimately resulting in the agent's decisions relying entirely on the DQN's output.

\subsection{Robust ANN-to-SNN Conversion and Error Alleviation}

As established in Section \ref{sec.problem}, directly applying traditional ANN-to-SNN conversion to RL agents introduces severe performance bottlenecks. 

The core principle of this conversion is mapping the ReLU activation of the original ANN, defined as $a^l=\text{ReLU}(W^la^{l-1}+b^l)$, to the equivalent firing rate of the converted SNN, defined as $\varphi^l(t)=\frac{1}{t}\sum_{t'=1}^t\theta^l(t')$. By accumulating the membrane potential over $t$ time-steps and assuming $v^l(0)=0$, this analytical relationship simplifies to:
\begin{equation}
    \varphi^l(t)=W^l\varphi^{l-1}(t)+b^l-\frac{v^l(t)}t\label{eq.conver}
\end{equation}

Equation \ref{eq.conver} shows theoretical equivalence with the ANN's ReLU activation. However, the residual membrane potential term, $\frac{v^{l}(t)}{t}$, introduces an unavoidable conversion error. At the ultra-low time-steps required for RMFS edge deployment, this error becomes critically magnified. To address the specific challenges of delayed spike emission and RL-SNN distribution mismatch caused by this error, we propose a two-stage error alleviation pipeline.

\begin{figure}[t]
\centering
\includegraphics[width=1\columnwidth]{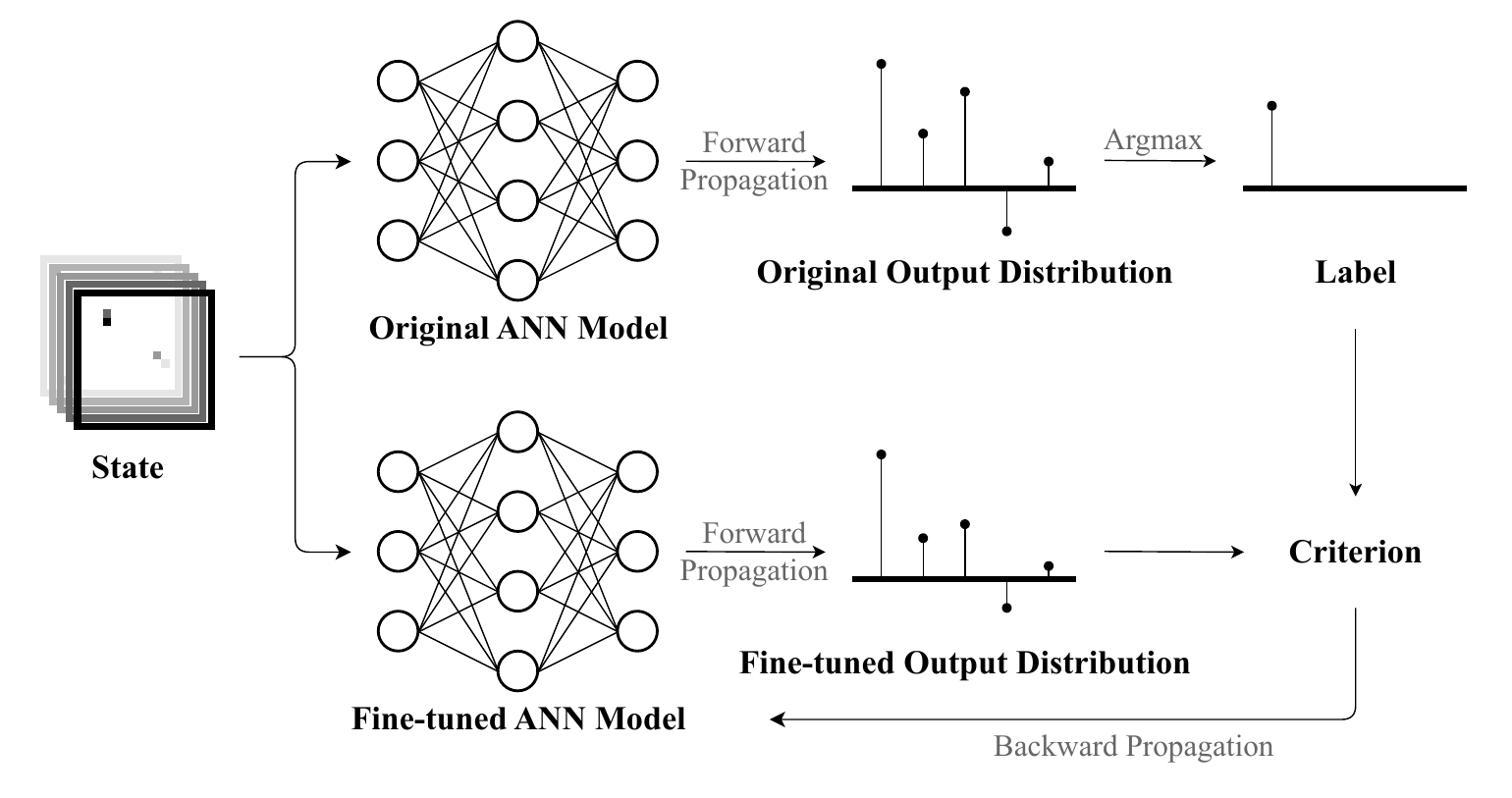}
\caption{The process of distillation before converting it to an SNN to alleviate conversion errors. The outputs of the original ANN are converted into a one-hot format using the argmax function, which serves as a label for training a new fine-tuned ANN. This method forces the fine-tuned ANN's outputs to approximate a one-hot distribution.}
\vspace{-0.4cm}
\label{fig.finetuning}
\end{figure}

\subsubsection{Parameter Scaling for Information Flow}

To mitigate delayed spike emission and accelerate information flow, we introduce a hardware-aware parameter scaling mechanism. Given the asynchronous processing nature of the neuromorphic chip's front-end, a lack of early spikes will stall the entire inference pipeline \cite{richter2023speck}. Therefore, we amplify the weights of the first layer by a scaling factor $k$, such that $W^1_{snn} = k \cdot W^1_{ann}$, which forces these neurons to emit spikes earlier and with higher frequency. This increased initial spiking activity acts as a critical trigger, guaranteeing rapid forward propagation of information through the deeper layers and ensuring the decision-making signal reaches the output under strict latency requirements (also see Table~\ref{table.ablation}). 

% It is important to acknowledge the inherent trade-off in this hard-label distillation scheme: collapsing the dense target distribution into one-hot pseudo-labels discards the fine-grained relative value information (i.e., the margin of utility between different valid actions). While this simplification could potentially degrade policy quality in highly complex, long-horizon tasks requiring subtle action ranking, it is highly effective for RMFS navigation. In dynamic warehouse environments, decisions primarily involve immediate directional choices and strict collision avoidance. For such highly contrasting and discrete decision spaces, widening the margin between the optimal action and sub-optimal alternatives is far more critical for mitigating low-timestep quantization noise than preserving precise continuous value differences.

Although this hard-label scheme inherently trades off fine-grained relative Q-value differences, it is highly advantageous for RMFS navigation. In such discrete spatial tasks, widening the decision margin of the optimal action to resist low-timestep quantization noise is far more critical than preserving precise continuous value rankings.

Retaining bias terms on event-driven neuromorphic chips requires extra synchronization overhead, which intrinsically forces a lowered operating frequency and increases static power consumption \cite{richter2023speck}. To avoid these hardware penalties and align with the chip's native architecture, we explicitly remove the bias components ($b^l=0$). This elimination sustains the low-latency performance demanded by real-time applications while ensuring optimal speed and energy efficiency without sacrificing the functional equivalence of the original ANN.
 
%Furthermore, to sustain this low-latency performance and align with the hardware's architecture, we explicitly remove the bias components ($b^l=0$). Retaining bias terms on such event-driven chips requires extra synchronization overhead, which intrinsically forces a lowered operating frequency and increases static power consumption \cite{richter2023speck}. Eliminating the bias avoids these hardware penalties entirely, ensuring optimal speed and energy efficiency without sacrificing the original ANN's functional equivalence.

\subsubsection{ANN-to-SNN Fine-tuning via Distillation}\label{sec.distill}

Existing ANN-to-SNN conversion methods typically focus on calibrating layer-wise activation distributions or elaborating SNN neuron dynamics to suppress intermediate errors~\cite{huang2025differential, li2021free, li2024error}. However, this indirect route is suboptimal for RL. If the true objective is a correct output decision, targeting it directly is far more efficient. Our analysis reveals that conversion mismatches primarily stem from output ambiguity rather than cumulative intermediate errors: valid actions often share similar Q-values, where minor conversion noise easily flips the final selection.

% Existing ANN-to-SNN conversion methods often focus on calibrating layer-wise activation distributions or complicating SNN neuron dynamics to reduce error. However, we found such layer-wise optimization to be unnecessary. Our analysis reveals that conversion mismatches primarily arise from output ambiguity—specifically, valid actions sharing very similar Q-values, where minor conversion noise easily alters the final prediction.

Targeting this output ambiguity, we reframe the pre-conversion fine-tuning phase as a classification-like knowledge distillation process (Figure \ref{fig.finetuning}). Using trajectories collected from the fully trained ANN, we duplicate the model into a frozen teacher and a trainable student. The teacher's Q-value outputs are passed through an $\text{argmax}$ function to generate hard, one-hot pseudo-labels, against which the student is fine-tuned via cross-entropy loss. Rather than preserving intermediate activation layers layer by layer, this formulation directly incentivizes the student ANN to widen the margin between the selected action and its alternatives, sharpening the output distribution into a robust, one-hot-like form while preserving the optimal policy. As a result, the network's decisions become inherently resilient to the unavoidable $\frac{v^{l}(t)}{t}$ quantization noise introduced during IF neuron conversion, cleanly bridging the continuous-discrete gap without the layer-wise optimization overhead (also see Table~\ref{table.ablation}).

\section{Experiment}
\subsection{Experiment Settings}
\subsubsection{Environment}

We developed an RMFS simulation system based on \cite{luo2023guiding} to conduct experiments in a typical warehouse scenario with multiple narrow aisles, several AGVs, and a complex pod arrangement. The environment measures $16\times 16$, holds 100 pods for transport, and features four picking stations, as shown in Figure \ref{fig.rmfs}.

In the RMFS, we tested the effectiveness of 1, 2, 4, and 8 AGVs working collaboratively to transport all pods. Each AGV starts from a designated position, with the system randomly generating task objectives involving the transportation of specified pods in three stages. New objectives are created after each task until all pods are successfully transported.

\subsubsection{Model Topology}
\begin{table}[t]
\centering
%\resizebox{.95\columnwidth}{!}{
\caption{Detailed architecture of the DQN}
\vspace{-0.3cm}
\begin{tabular}{ccc}
        \hline
    Layer& Components & Shape of feature \\
        \hline
    1 & $3\times 3$ Conv & $32\times16\times16$  \\
    2 & $3\times 3$ Conv & $32\times16\times16$\\
    3 & $3\times 3$ Conv + AvgPool & $32\times8\times8$  \\
    4 & $3\times 3$ Conv & $64\times8\times8$\\
    5 & $3\times 3$ Conv + AvgPool & $32\times4\times4$\\
    6 & $3\times 3$ Conv &$128\times4\times4$\\
    7 & $3\times 3$ Conv + AvgPool& $32\times1\times1$\\
    8 & $1\times 1$ Conv &$512\times1\times1$\\
    9 & $1\times 1$ Conv & $5(4)\times1\times1$\\
        \hline
\end{tabular}
\label{table.arch}
\end{table}

The DQN we employed is designed based on a convolutional neural network, consisting of nine convolutional layers comprised of $3\times3$ and $1\times1$ convolutional cores. To meet hardware deployment requirements, we utilized average pooling (AvgPool). The detailed architecture of the DQN is shown in Table \ref{table.arch}. Each convolutional layer is followed by a ReLU activation, and the shape of the last layer's feature map varies based on the number of agents, as described in Section \ref{sec.problem}.

\subsubsection{Agent Deployment}
After training the ANN and converting it to an SNN, we deployed the SNN in the RMFS for AGV pathfinding tasks. %Given the limited capabilities of SNNs on neuromorphic chips,
We implemented an action detection mechanism for the SNN agents, which checks for potential collisions before executing an action; if a collision is highly likely, the action is changed to stop. While this external rule-based safety layer is a standard engineering practice to guarantee absolute safety in physical robotic systems, our framework relies on the inherently high Conversion Rate $CR$ of the fine-tuned SNN to ensure accurate primary decisions, thereby minimizing the frequency of triggering these safety overrides.

\subsection{Effect of Collision Limit} 
\begin{figure*}[t]
\centering
\includegraphics[width=0.9\textwidth]{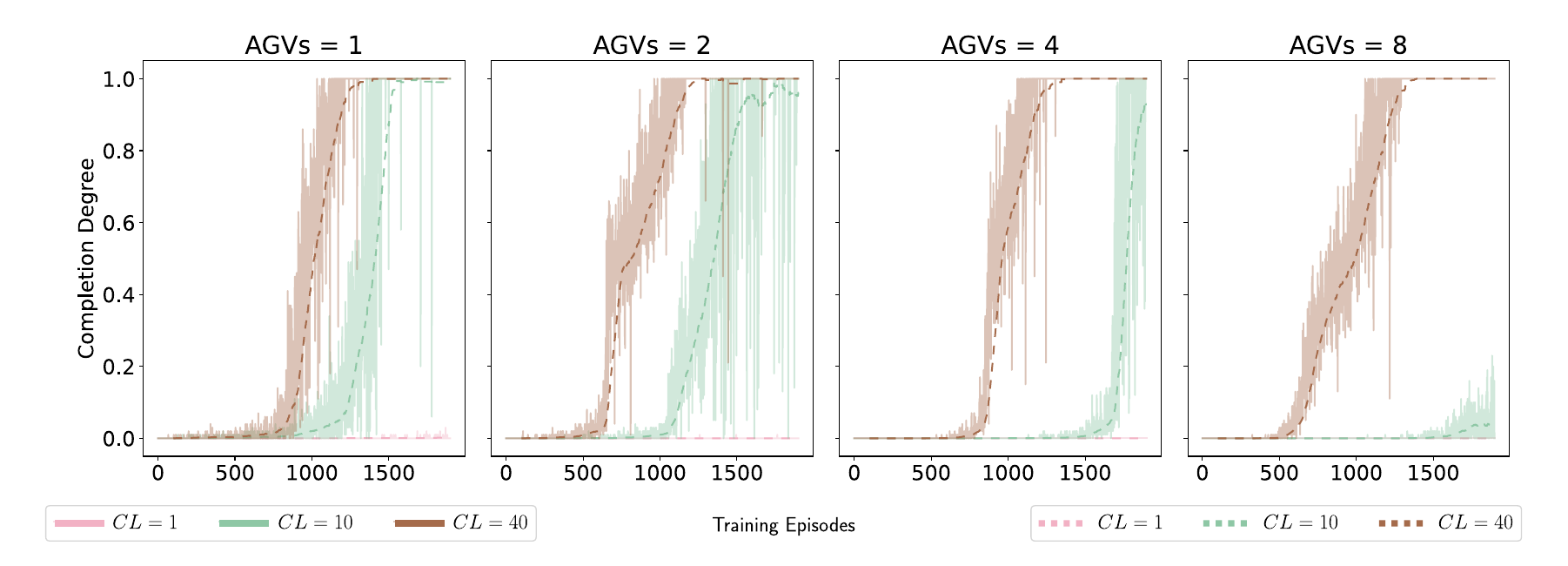}
\vspace{-0.5cm}
\caption{The impact of different $CL$ on the DQN's training process for various numbers of AGVs. The figures show the completion degree, which is the ratio of successfully transported pods. The dashed lines represent the smoothing of the completion degree and action length curves.}
\vspace{-0.5cm}
\label{fig.collision}
\end{figure*}

% Figure \ref{fig.collision} illustrates the effect of varying collision limits on the training outcomes during the DQN training process. The term completion degree refers to the ratio of the number of successfully transported pods to the total number of pods that need to be moved within the RMFS environment. A completion degree of 1 indicates that the AGVs have accomplished a complete cycle of the transport task. As the collision limit increases (from $CL=1$ to $CL=40$), the completion degree curves show more stability and efficiency, especially with more AGVs.

% As illustrated, a higher $CL$ contributes to the stability and efficiency of DQN training, particularly when the number of AGVs is large. The advantages of our collision-allowing training method become more pronounced under these conditions. Subsequent steps utilize models trained with a setting of $CL=40$.

Figure \ref{fig.collision} shows the impact of collision limits on DQN training. Here, completion degree represents the ratio of successfully transported pods to the total task. As $CL$ increases from 1 to 40, the training curves exhibit enhanced stability and efficiency, particularly in high-density AGV scenarios. These results highlight the advantages of our collision-allowing training method, leading us to adopt CL=40 for subsequent experiments.

\subsection{DQN Alternatives}\label{sec.alt}
\begin{figure}[t]
\centering
\includegraphics[width=0.99\columnwidth]{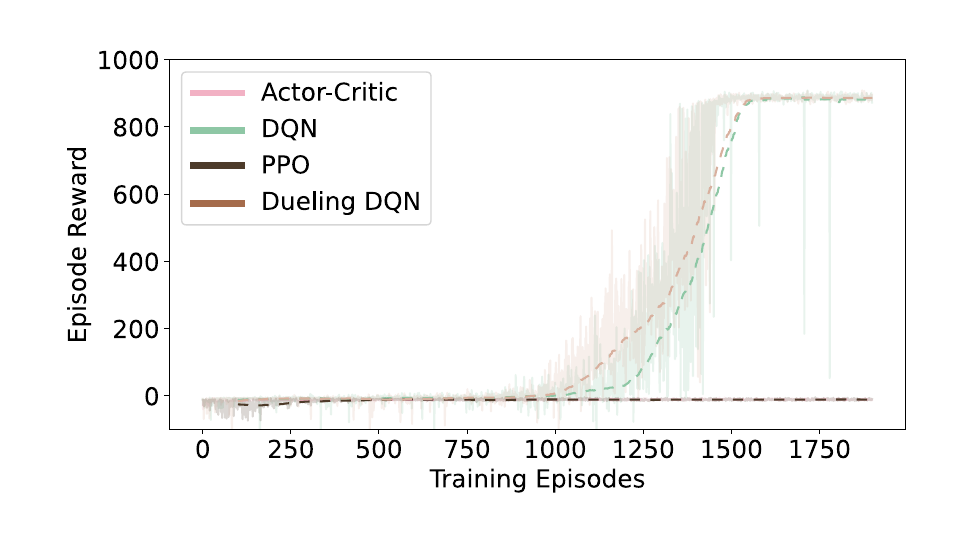}
\vspace{-1cm}
\caption{Comparison of different standard RL baselines such as Dueling DQN and PPO. The dashed line represents the smoothing of the episode reward.}
\vspace{-0.5cm}
\label{fig.alt}
\end{figure}

We benchmark our RL policy against three established baselines, namely Dueling DQN, Proximal Policy Optimization (PPO), and standard Actor-Critic~\cite{wang2016dueling,schulman2017proximal,konda1999actor}. All algorithms were trained within a single-AGV scenario in the same RMFS environment, using identical sparse state representations and substantially consistent model topology, with the collision limit set to 10.

As illustrated in Figure \ref{fig.alt}, the experimental results reveal a stark contrast in learning capabilities. The policy-based methods, PPO and Actor-Critic, completely failed to perform pathfinding throughout the entire training process. Conversely, value-based methods, DQN and Dueling DQN, successfully converged to high episode rewards. While Dueling DQN exhibited slightly faster initial convergence, both algorithms eventually reached comparable final pathfinding performance.

Although Dueling DQN learns slightly faster, its network architecture introduces additional complexity and potential accuracy degradation during ANN-to-SNN conversion. DQN not only reliably solves the RMFS pathfinding task but also features a straightforward architecture, serving as the most optimized and robust foundation for our low-loss ANN-to-SNN conversion pipeline.

\subsection{Effect of ANN-to-SNN Error Alleviation}
\begin{table*}[t]
\centering
\caption{Effect of ANN-to-SNN Error Alleviation Method}
\vspace{-0.2cm}
%\resizebox{.95\columnwidth}{!}{
\begin{tabular}{cccccc|cccc}
        \hline
        
   \multirow{2}{*}{AGVs} & \multirow{2}{*}{Distillation}
    &\multicolumn{4}{c}{$CR$ (\%)}&\multicolumn{4}{c}{Completion Degree (\%) / Action Length}\\
    \cline{3-10}
    &&$k=1$&$k=3$&$k=5$&$k=7$&$k=1$&$k=3$&$k=5$&$k=7$\\
        \hline
   \multirow{2}{*}{1} 
   &\XSolidBrush&27.04&40.65&55.21&64.28&0.00 / --&0.00 / --&0.00 / --&0.05 / --\\
   &\Checkmark &27.91&54.45&94.51&\textbf{99.60}&0.00 / --&45.65 / --&100.00 / 3609.44&\textbf{100.00 / 3609.38}\\
   \multirow{2}{*}{2} 
   &\XSolidBrush&28.87&40.95&55.06&65.20&0.00 / --&0.02 / --&0.70 / --&12.00 / --\\
   &\Checkmark &30.36&78.58&97.98&\textbf{99.44}&0.00 / --&91.71 / 4049.26&\textbf{97.94 / 3941.04}&93.17 / 3940.52\\
   \multirow{2}{*}{4} 
   &\XSolidBrush&26.50&28.93&38.80&48.77&0.00 / --&0.00 / --&0.00 / --&0.49 / --\\
   &\Checkmark &28.30&65.57&92.85&\textbf{97.87}&0.00 / --&\textbf{99.98 / 3963.54}&99.96 / 4012.87&99.94 / 4082.93\\
   \multirow{2}{*}{8} 
   &\XSolidBrush&28.26&40.12&55.21&65.57&0.00 / --&0.51 / --&97.79 / 17999.70&99.66 / 6462.03\\
   &\Checkmark &28.07&71.32&95.13&\textbf{97.84}&0.00 / --&99.33 / 6303.03&99.67 / 5631.44&\textbf{99.64 / 5416.32}\\
        \hline
\end{tabular}
\label{table.ablation}
\end{table*}

We evaluate the performance metrics of AGVs' pathfinding under various configurations to demonstrate the effect of our ANN-to-SNN error alleviation method. The results are summarized in Table \ref{table.ablation}, which presents the $CR$ values, completion degree, and action lengths for different numbers of AGVs (1, 2, 4, and 8) under varying scaling factors ($k=1$, $k=3$, $k=5$, and $k=7$) with and without fine-tuning. $CR$ is the conversion rate indicating the accuracy of producing the same output for the same input between the converted SNN and the original ANN. Notably, the time-step of SNN is set to 4 for efficient inference.

\begin{figure}[t]
\centering
\includegraphics[width=0.99\columnwidth]{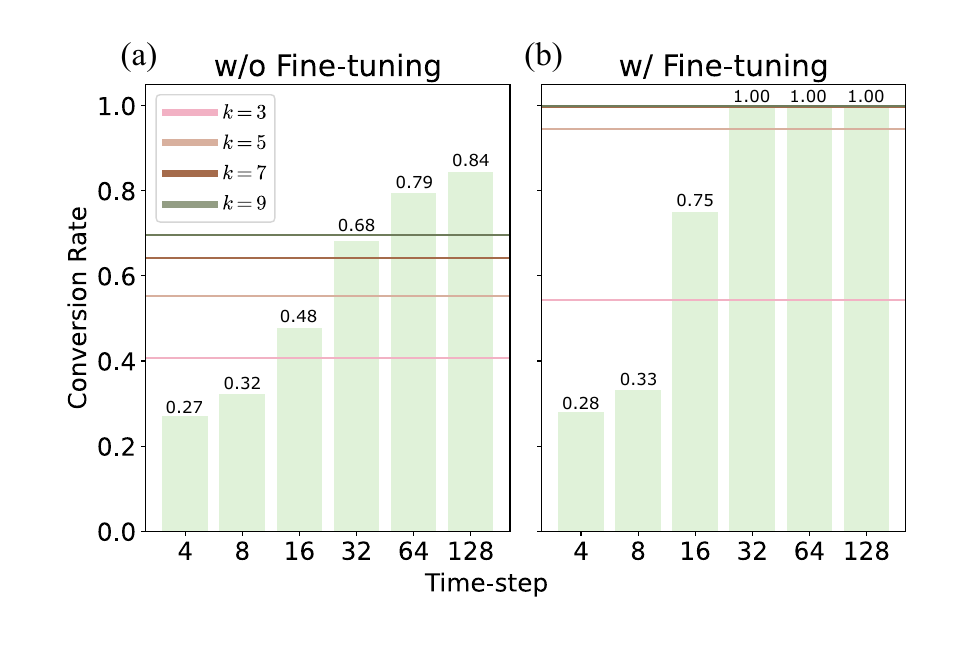}
\vspace{-1cm}
\caption{The ANN-to-SNN conversion rate $CR$ under various time-step settings. (a) shows the performance of an SNN converted directly from an ANN without fine-tuning, while (b) shows the performance with fine-tuning. The colored horizontal lines represent the $CR$ of ANN-to-SNN with different scaling factors and time-step set to 4.}
\vspace{-0.5cm}
\label{fig.cr}
\end{figure}

We evaluated the conversion rates ($CR$) across various configurations. As illustrated in Figure \ref{fig.cr}, a directly converted SNN without fine-tuning or scaling fails to accurately capture the original policy, peaking at a $CR$ of 0.84 despite utilizing 128 time-steps, which leads to increased inference latency and energy consumption. Conversely, our fine-tuned ANN-to-SNN substantially accelerates this process, achieving a perfect $CR$ of 1.00 within 32 time-steps. The integration of parameter scaling further minimizes the temporal requirement; for instance, applying a scaling factor of $CR$ to the fine-tuned model sustains a lossless $CR$ of 1.00 at an ultra-low 4 time-steps. These results demonstrate that combining parameter scaling with ANN-to-SNN distillation guarantees optimal efficiency and decision fidelity under strict latency constraints.

Furthermore, within the context of our framework, a marginal drop in $CR$ does not necessarily precipitate severe performance degradation. This systemic robustness is largely attributed to the action detection mechanism integrated during deployment. Should quantization errors occasionally produce a suboptimal or collision-prone action, the safety filter preemptively overrides it with a 'stop' command. Thus, instead of failing the episode, the agent safely absorbs the single-step error and replans using updated state observations in the subsequent time-step. Crucially, since our fine-tuned distillation ensures a near-lossless $CR$ (e.g., 0.99 to 1.00), interventions by this safety override are exceedingly rare. The converted SNN inherently retains the pathfinding mastery of the original policy, relegating the closed-loop safety filter to a mere pragmatic safeguard against residual quantization noise.

\subsection{Hardware Deployment and Performance}\label{sec.hard}
\begin{table*}[t]
\centering
\caption{Hardware Deployment and Performance}
\vspace{-0.2cm}
\resizebox{.99\textwidth}{!}{
\begin{tabular}{c|ccc|ccc|ccc|ccc}
        \hline
    AGVs& \multicolumn{3}{c|}{1} & \multicolumn{3}{c|}{2} & \multicolumn{3}{c|}{4}&\multicolumn{3}{c}{8}\\
        \hline
   Model & ANN &SNN&SNN & ANN & SNN&SNN  & ANN & SNN&SNN & ANN &SNN&SNN \\
        % \hline
   Device&GPU&GPU&Speck&GPU&GPU&Speck&GPU&GPU&Speck&GPU&GPU&Speck\\
   % \hline
   % Energy ($\mu \text{J}$)&37.4&8.5&\textbf{1.46}&37.7&20.9&\textbf{1.94}&37.7&15.8&\textbf{1.83}&37.7&22.9&\textbf{2.87}\\
   % Latency ($\text{ms}$)&0.19&3.23&\textbf{$<$0.1}&0.20&3.24&\textbf{$<$0.1}&0.20&3.26&\textbf{$<$0.1}&0.19&3.28&\textbf{$<$0.1}\\
   %      \hline
   \hline
   Energy ($ \text{mJ}$)&10.04&188.47&$\mathbf{0.89\times10^{-3}}$&10.73&184.74&$\mathbf{1.53\times10^{-3}}$&10.42&186.85&$\mathbf{1.21\times10^{-3}}$&10.26&182.80& $\mathbf{1.95\times10^{-3}}$\\
   Latency ($\text{ms}$)&0.13&2.42&\textbf{0.061}&0.14&2.44&\textbf{0.079}&0.13&2.41&\textbf{0.066}&0.13&2.44&\textbf{0.068}\\
        \hline
\end{tabular}
}
\label{table.power}
\end{table*}
We profile the energy consumption and inference latency of the ANN and its corresponding SNN across different hardware platforms to assess the physical inference viability of the SDQN-RMFS framework at the edge. For fair comparison, we first deployed the SNN on a GPU and recorded its trajectory data within the RMFS to create a dataset for energy measurement. Subsequently, the original ANN and the converted SNN were deployed on an NVIDIA RTX 4090 GPU and the specialized neuromorphic chip, Speck. By sequentially inputting the state data from the trajectory dataset, we measured the operational performance of the models using the respective profiling tools. For the converted SNN, the number of time-steps was set to 4 with a scaling factor of 7. The comparative results across various multi-AGV scenarios are detailed in Table \ref{table.power}.

\subsubsection{Energy Efficiency}
As shown in Table \ref{table.power}, the standard ANN deployed on an RTX 4090 GPU consumes approximately 10.04 to 10.73 mJ per decision. Notably, when simulating the SNN on the same GPU, the energy consumption surges to between 182.80 and 188.47 mJ due to the inherent inefficiencies of the Von Neumann architecture, which requires extremely high-overhead serial memory reads for multiple spiking time-steps. This indicates that traditional accelerators are fundamentally ill-suited to the sparse dynamic characteristics of SNNs.

% However, the immense advantages of the spiking mechanism are immediately unleashed when the SNN is deployed on the neuromorphic chip. The SNN on the Speck achieves an ultra-low energy consumption of only $0.89\times10^{-3}$ to $1.95\times10^{-3}$ mJ. Compared to the ANN baseline on the GPU, this represents a breakthrough energy saving of several thousand times, proving that the framework is well-equipped for the long-term, continuous operational requirements of edge devices.

However, the immense advantages of the spiking mechanism are immediately unleashed when the SNN is deployed on the neuromorphic chip. The SNN on the Speck achieves an ultra-low energy consumption of only $0.89\times10^{-3}$ to $1.95\times10^{-3}$ mJ. Compared to the GPU-ANN baseline, this represents a breakthrough energy reduction of up to 11,281$\times$. Such ultra-low-power pathfinding ensures that continuous, large-scale multi-AGV operations in RMFS are not bottlenecked by computational energy constraints.

\subsubsection{Real-Time Inference Latency}
% The latency data in Table \ref{table.power} confirms the critical importance of hardware-software alignment. On the RTX 4090, the original ANN leverages its immense tensor computing power to achieve an inference time of only 0.13 to 0.14 ms per decision. However, when the GPU attempts to process the SNN, the iterative simulation of multiple time-steps causes the latency to deteriorate significantly to over 2.41 ms. In contrast, the Speck is capable of processing binary spike events natively and asynchronously, further compressing inference latency to 0.061 to 0.079 ms. Remarkably, SDQN-RMFS not only reduces energy consumption by five orders of magnitude, but it actually halves the latency of a powerful GPU while running on a strictly power-constrained edge chip. This breakthrough fundamentally unlocks the capability for highly reactive, large-scale multi-AGV coordination in dynamic warehouse environments.

Table \ref{table.power} confirms the critical importance of hardware-software alignment through the latency data. On the RTX 4090, the original ANN leverages its immense tensor computing power to achieve an inference time of only 0.13 to 0.14 ms per decision. However, when the GPU attempts to process the SNN, the iterative simulation of multiple time-steps causes the latency to deteriorate significantly to over 2.41 ms. In contrast, the Speck chip is capable of processing binary spike events natively and asynchronously, further compressing inference latency to 0.061 to 0.079 ms. Remarkably, SDQN-RMFS not only reduces energy consumption by five orders of magnitude but actually halves the latency of a powerful GPU while running on a strictly power-constrained edge chip. This breakthrough fundamentally unlocks the capability for highly reactive, large-scale multi-AGV coordination in dynamic warehouse environments.

% Relative to conventional ANN inference on GPU, our neuromorphic deployment reduces energy consumption by orders of magnitude while also achieving significantly lower inference latency.

% In large-scale RMFS deployments, ultra-low inference latency is a strict operational prerequisite rather than a mere optimization. When hundreds of AGVs navigate simultaneously through narrow, dynamically changing aisles, even millisecond delays in decision-making can trigger cascading deadlocks or collisions. While state-of-the-art ANN policies deployed on high-end desktop accelerators like the RTX 4090 can achieve an inference time of 0.13 to 0.14 ms per decision, deploying such power-hungry hardware on resource-constrained AGVs is physically impossible. On realistic edge devices, this baseline latency would degrade exponentially, crippling real-time responsiveness. Furthermore, when the GPU attempts to process the SNN, iterative simulation of multiple time-steps causes the latency to deteriorate significantly to over 2.41 ms.

% This is where the SDQN-RMFS framework demonstrates its disruptive capability. By perfectly aligning our optimized SNN with the asynchronous architecture of the Speck chip, the hardware processes binary spike events natively and instantaneously. This compresses the inference latency to an astonishing 0.061 to 0.079 ms. 
\section{Conclusions}
This study presented SDQN-RMFS, a synergistic framework integrating DQNs with SNNs for energy-efficient multi-AGV pathfinding. This innovative framework tackles the challenges of traditional algorithms by combining the low-power, event-driven characteristics of SNNs with the optimal policy learning capabilities of DQNs.

During training, a collision-allowing strategy is employed to pretrain the ANN, preventing premature termination caused by frequent collisions and significantly improving training stability in multi-AGV scenarios. The trained ANN is subsequently converted into an SNN through a fine-tuned conversion method incorporating parameter scaling and distillation, which minimizes time-steps and accuracy loss while preserving performance comparable to the original ANN.

A key contribution of this work is the successful physical offline deployment of the trained SNN onto a neuromorphic chip for inference. By comprehensively addressing the conversion mismatch and utilizing hardware-aware parameter scaling, this work demonstrates a critical and empirically validated step toward practical applicability in energy-constrained, closed-loop mobile robotic systems.

\bibliographystyle{IEEEtran} 
\bibliography{reference,IEEEabrv}

\addtolength{\textheight}{-12cm}   % This command serves to balance the column lengths

\end{document}